\documentclass{article}
\usepackage{spconf,amsmath,graphicx}

\usepackage{subfig}
\usepackage{color}
\usepackage{enumitem}
\usepackage{amssymb}
\setlist{nosep, leftmargin=14pt}
\usepackage{amsfonts}
\usepackage{xcolor}
\usepackage{soul}

\AtBeginDocument{
  \hypersetup{
    urlcolor=red,
    citecolor=magenta,
    linkcolor=blue,
  }%
}
\usepackage[colorlinks=true,]{hyperref}

\usepackage{mwe} 

\usepackage{algorithm}
\usepackage{algpseudocode}

\title{Reconstructing the Normal and Shape at Specularities in Endoscopy}
\name{Karim Makki and Adrien Bartoli\thanks{This work was funded by the FET-Open grant 863146 Endomapper.}}
\address{EnCoV-Institut Pascal, CNRS/Université Clermont-Auvergne, Clermont-Ferrand 63000, France}

\begin{document}
%
\maketitle
\begin{abstract}
Specularities are numerous in endoscopic images. 
They occur as many white small elliptic spots, which are generally ruled out as nuisance in image analysis and computer vision methods.
Instead, we propose to use specularities as cues for 3D perception.
Specifically, we propose a new method to reconstruct, at each specularity, the observed tissue's normal direction ({\em i.e.}, its orientation) and shape ({\em i.e.}, its curvature) from a single image.
We show results on simulated and real interventional images.
\end{abstract}
\section{Introduction}
\label{sec:intro}
Specularities on smooth reflecting surfaces depend on the surface's local orientation and curvature, on the light source and on the camera position~\cite[Chapter 2]{forsyth2002computer}. In the general setting, all these contain far too many unknowns to allow one to exploit specularities effectively in 3D reconstruction. 
However, endoscopy is a particular case where the {\em light source and camera form a collocated rig}. 
Specularities are numerous in endoscopy, where the moist tissue causes specular reflection.
However, they are generally considered as artifacts in endoscopic 3D reconstruction methods, both in classical methods such as Shape-from-Shading (SfS)~\cite{ciuti2012intra} and learning-based methods such as direct depth estimation~\cite{liu2023self}. 
More recently, it has been understood that specularities could form useful cues for 3D reconstruction.
The study of~\cite{daher2023cyclesttn} has shown that training a learning-based reconstruction model from synthetic data which can eventually deal with real data requires one to generate specularities.
Concretely, we have recently shown that the surface normal could be reconstructed at each specularity~\cite{makki2023normal, makki2023elliptical}, with a two-fold ambiguity.
The reconstruction method uses the {\em specular isophote}, which is the curve enclosing the specularity in the image.
Based on the assumption that the surface underlying the specularity is flat, we have used a mathematical model to show that the isophote is elliptic.
This approach has advantages: the method runs fast and can process full high image resolution, in contrast to the learning-based approach which is currently limited to lower image resolution to keep runtime reasonable (the available datasets such as the EndoSLAM~\cite{ozyoruk2021endoslam} and UCL~\cite{rau2019implicit} ones are at lower resolution).
However, it also has strong limitations.
First, the reconstruction is up to a two-fold ambiguity: only one of the two directions found at each point is the correct normal.
Second, the local planarity assumption is unreasonable, as specularities typically appear near highly curved surface patches.
{\em Is an ambiguous surface normal all one can reconstruct from a specularity?}

We propose to move a step ahead by using a more advanced local surface model.
Specifically, we leave the restrictive flat surface to adopt a curved surface model.
We show that with this model, the specular isophote is still elliptical in the image, albeit that its eccentricity is not just due to surface orientation as in past work, but to both surface orientation and curvature.
We show that one can reconstruct the surface normal from just the {\em specularity's Brightest Point} (BP) and the surface curvature from the elliptical isophote.
Specifically, we show that the principal curvature directions (the directions of minimum and maximum curvatures on the tangent plane) and the curvature ratio can be reconstructed.
Therefore, {\em the specular isophote contains much more information} than one could think from past work.
We propose an automatic reconstruction method, which, from an endoscopic image and the camera's intrinsic parameters, detects the specularities and reconstructs the local surface orientation and curvature ratio at each specularity.
This reconstruction can be used to implement downstream tasks such as computer-aided navigation, to bootstrap other reconstruction methods such as SfS and to train depth estimation networks.
We evaluate the method on realistic simulated colonoscopic data from Blender and surgical laparoscopy images.

\section{Theory and methods}
\label{sec:methods}
\subsection{Mathematical Model}
\label{sssec:model}
We consider a calibrated perspective camera model with intrinsic parameters in matrix $\mathbf{K} \in \mathbb{R}^{3 \times 3}$ and a point light source near the camera centre located at the origin. We consider a smooth specular surface and assume that the brightness reﬂected along the sightline direction $V$ is a function of $V \cdot P$ , where $P$ is the {\em specular direction}, also known as the direction of a perfect specular reflection. This assumption is commonly used in reflection models such as Blinn-Phong~\cite{blinn1977models} and the physics-based BSDF model popularised by Disney~\cite{burley2012physically}.
In these models, each specularity has a BP, approximately located at its centre, where the reflected light energy is the highest.
Let $L$ be the lighting direction; it is known that, from the fact that the vectors $L$, $V$ and $P$ have unit length~\cite{forsyth2002computer,blinn1977models,burley2012physically}, the half-angle direction $H = (L + V) / 2$ is collinear with the surface normal $\mathbf{N}$ at the specularity's BP. In endoscopy, as we have $L \approx  V$, the surface normal is thus collinear with the sightline $V$.

\begin{figure}[!h]
\centering
\includegraphics[width=.3\textwidth]{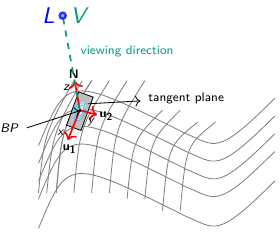}\quad
\includegraphics[width=.16\textwidth]{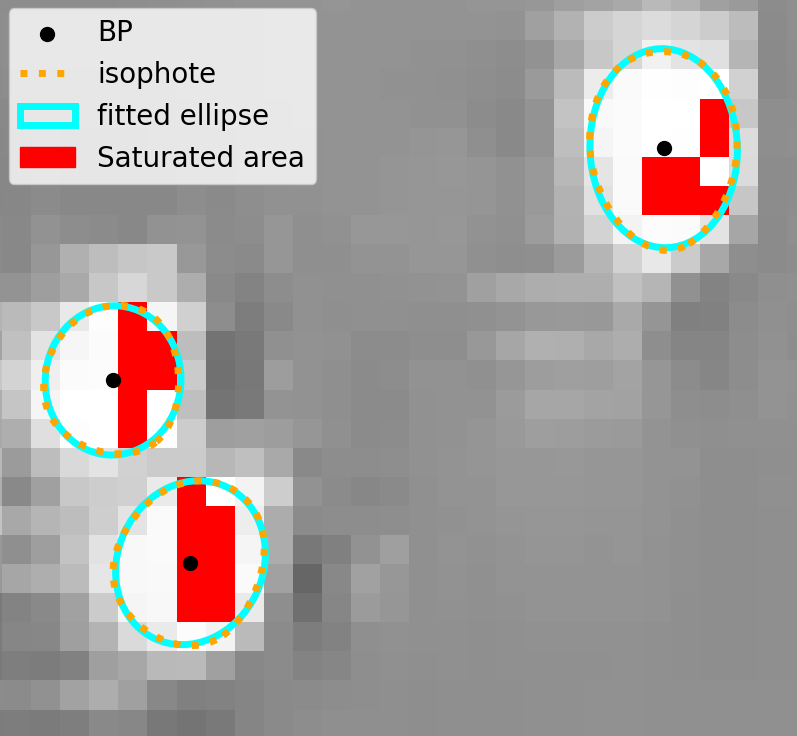}
\caption{\label{fig:fig1} \textbf{Left:} local coordinate system at the BP oriented using the normal and principal directions. The light and camera are collocated. Elliptical specular isophotes are shown in cyan. \textbf{Right:} an example of three neighbouring surface patches from a real endoscopic image, showing the difficulty to determine the BP directly from image intensities.}
\end{figure}

We model the observed tissue as a collection of 3D surface patches.
We model each patch by three components, as shown in figure~\ref{fig:fig1} (left): 1) its {\em normal direction} $\mathbf{N}$, 2) its {\em tangent plane}, with basis vectors $\mathbf{u}_1, \mathbf{u}_2$ chosen along the patch's principal directions ({\em i.e.}, the directions of minimal and maximal curvature), and 3) its {\em curvature ratio}, which is the ratio of the minimal to the maximal curvature.
We instantiate one patch per specularity.
We assume the specularity to conform to the elliptic model, which is widely applicable in endoscopic images, where they occur as small isolated white blobs~\cite{forsyth2002computer, makki2023normal}.



\subsection{Reconstruction Method}
\label{sssec:reconstruction} 
Our reconstruction method has three main steps given as Algorithm~\ref{alg_isbi}.
The first step, which is as in~\cite{makki2023elliptical}, uses a learning-based specularity detector and ellipse fitting to retrieve an isophote for each specularity.
We perform a simple ellipticity test based on the fitting residual.
The second and third steps are geometry-based reconstruction given directly below.

\subsubsection{Sightline-based Normal Reconstruction}
\label{sssec:Sightline-based} 
We exploit the property that, at the BP, the surface normal is collinear with the sightline.
We approximate the BP by the elliptic isophote's centre $(x_0,y_0)$ in image pixel coordinates, as shown in figure~\ref{fig:fig1} (right).
The sightline is trivially found using the camera model, giving the surface normal direction  $\mathbf{N}\in S^2$ as $\mathbf{N} \propto \mathbf{K}^{-1} (x_0,y_0,1)^\top$, where $S^2$ is the set of direction vectors in 3-space. Obviously, the reconstruction accuracy directly depends on the accuracy of the  BP estimate. However, determining the exact localtion of the BP is challenging because of camera saturation at specularities. As seen in the right side of figure~\ref{fig:fig1}, all the red pixels have the maximal intensity for the selected surface patch. In contrast, it is much easier to detect an isophote whose centre and the unknown BP are nearly coincident. 

\subsubsection{Ellipse-based Shape Reconstruction}
\label{algorithm} 

We represent a specular isophote ellipse by its standard  $3 \times 3$ real symmetric matrix form $C$.\\
This satisfies $(x,y,1)  C  (x,y,1)^\top=0$ for the image pixel coordinates $(x,y)$ of an isophote point.
Next, we map the conic $C$ to the normalized image plane\footnote{The origin is at the principal point and the distances are measured in units of focal length.} as $C' = \mathbf{K}^\top C \mathbf{K}$. Note that $C'$, in its turn, is a $3 \times 3$ real symmetric matrix. It has therefore the following properties: 1) always diagonalizable; 2) it has real eigenvalues; and 3) its eigenvectors are orthogonal.
We then diagonalise $C'$ as $C' = \mathbf{R}^\top  diag(\lambda_1, \lambda_2, \lambda_3) \mathbf{R}$, where $\{\lambda_1, \lambda_2, \lambda_3\}$ are the eigenvalues sorted in ascending order. The corresponding eigenvector set $\{\vec{v_1}, \vec{v_2}, \vec{v_3}\}$ forms an orthonormal basis of $\mathbb{R}^3$. In fact, $\mathbf{R} = (\vec{v_1}, \vec{v_2}, \vec{v_3})$ is a 3D rotation of the standard camera coordinate system to the so-called \textit{eigenvector-frame}. It aligns the principal directions on the tangent plane at the BP with the ellipse's major and minor axes, with the normal direction $\mathbf{N}$ being aligned with the cross product $\vec{v_3} = \vec{v_1} \times \vec{v_2}$. The principal directions on the tangent plane are therefore given by $\mathbf{u}_1, \mathbf{u}_2 = \vec{v_1}, \vec{v_2}$. The ratio of principal curvatures $|\lambda_1/\lambda_2|$ can be derived from the ellipse's eccentricity defined as $\epsilon= \sqrt{1 - (\lambda_1/\lambda_2)^2}$.

\begin{algorithm}
\caption{Single-image specular normals and shapes reconstruction}\label{alg_isbi}
\begin{algorithmic}
\Require Image $I$, camera intrinsics $\mathbf{K}$
\Ensure Sets of Brightest Points (BP), normals $\mathbf{N}$, principal directions $\mathbf{u}_1,\mathbf{u}_2$, curvature ratios $\epsilon$
\State - Use the neural network to segment the specular mask $M$ from  $I$
\Comment{\cite{makki2023elliptical}}
\For  {each connected component $c \in M$} 
\State  - find the specular isophote as the outer boundary of $c$
\State  
\parbox[t]{\dimexpr\linewidth-\algorithmicindent}{%
       - smooth the isophote using a cubic B-spline
        }
        \State
\parbox[t]{\dimexpr\linewidth-\algorithmicindent}{%
  - fit an ellipse to the isophote, with matrix representation $C$ and fitting residual $r$}
\State
\If{$r \leq t$}  \Comment{ellipticity test with threshold $t$} 

\State - find the BP $(x_0,y_0)$ as the ellipse centre
\State
\parbox[t]{\dimexpr\linewidth-\algorithmicindent}{%
        - compute the normal direction $\mathbf{N} = \mathbf{K}^{-1} (x_0,y_0,1)^\top$ %
        }
\State - normalise it as $\mathbf{N} = \mathbf{N}/  \|  \mathbf{N} \|$ 

\State 
\parbox[t]{\dimexpr\linewidth-\algorithmicindent}{%
        - transfer the ellipse to the normalised image plane as $C' \gets \mathbf{K}^\top C \mathbf{K}$
        }

\State   \parbox[t]{\dimexpr\linewidth-\algorithmicindent}{%
       - retrieve the ellipse axes by eigen-decomposition as $C' = \mathbf{R}^\top  diag(\lambda_1, \lambda_2, \lambda_3) \mathbf{R}$, with eigenvalues $\lambda_1 \leq \lambda_2 \leq \lambda_3$ and eigenvectors $\mathbf{R} = (\vec{v_1}, \vec{v_2}, \vec{v_3})$} 

\State 
\parbox[t]{\dimexpr\linewidth-\algorithmicindent}{%
        - determine the principal directions on the tangent plane as $\mathbf{u}_1, \mathbf{u}_2 = \vec{v_1}, \vec{v_2}$ %
        }

\State 
\parbox[t]{\dimexpr\linewidth-\algorithmicindent}{%
        - compute the curvature ratio from the ellipse eccentricity $\epsilon$%
        }

\EndIf 
\EndFor
\end{algorithmic}
\end{algorithm}


\section{Experimental results}
\label{sec:results}

\subsection{Synthetic Colonoscopic Data}
\label{subsec:synthetic}

\subsubsection{Model}
\label{subsubsec:Illumination}
We use the modelling and rendering software suite Blender to synthesise images using a wide-angle perspective camera of 10~mm focal length.
We use a spotlight located at the camera centre with a radius of 0.1~mm and projected to a 120$^{\circ}$ cone. 
We generate a colon-like deforming surface with free-form deformation~\cite{joshi2007harmonic}.
The generated surface is represented by a triangular mesh with 502,502 vertices and 1,002,000 faces.
We deform it over 250 frames. 
The surface is imparted with a strong `clearcoat' to boost specular reflections. 
We use the anisotropic principled BSDF model~\cite{burley2012physically}, mimicking real-world shading and specularities. 


\subsubsection{Normal Reconstruction}
\label{subsubsec:Normal reconstruction}
Our method found 1,500 elliptic specularities.
The reconstructed normals using the proposed sightline based method are compared with the true mesh normals, at mesh vertices corresponding to the set of detected BPs in the image. The histogram of errors in normal estimates is shown in figure~\ref{fig:3}. 

\subsubsection{Shape Reconstruction}
\label{subsubsec:principal directions}
\begin{figure}[!t]
\centering
\includegraphics[width=0.5\textwidth]{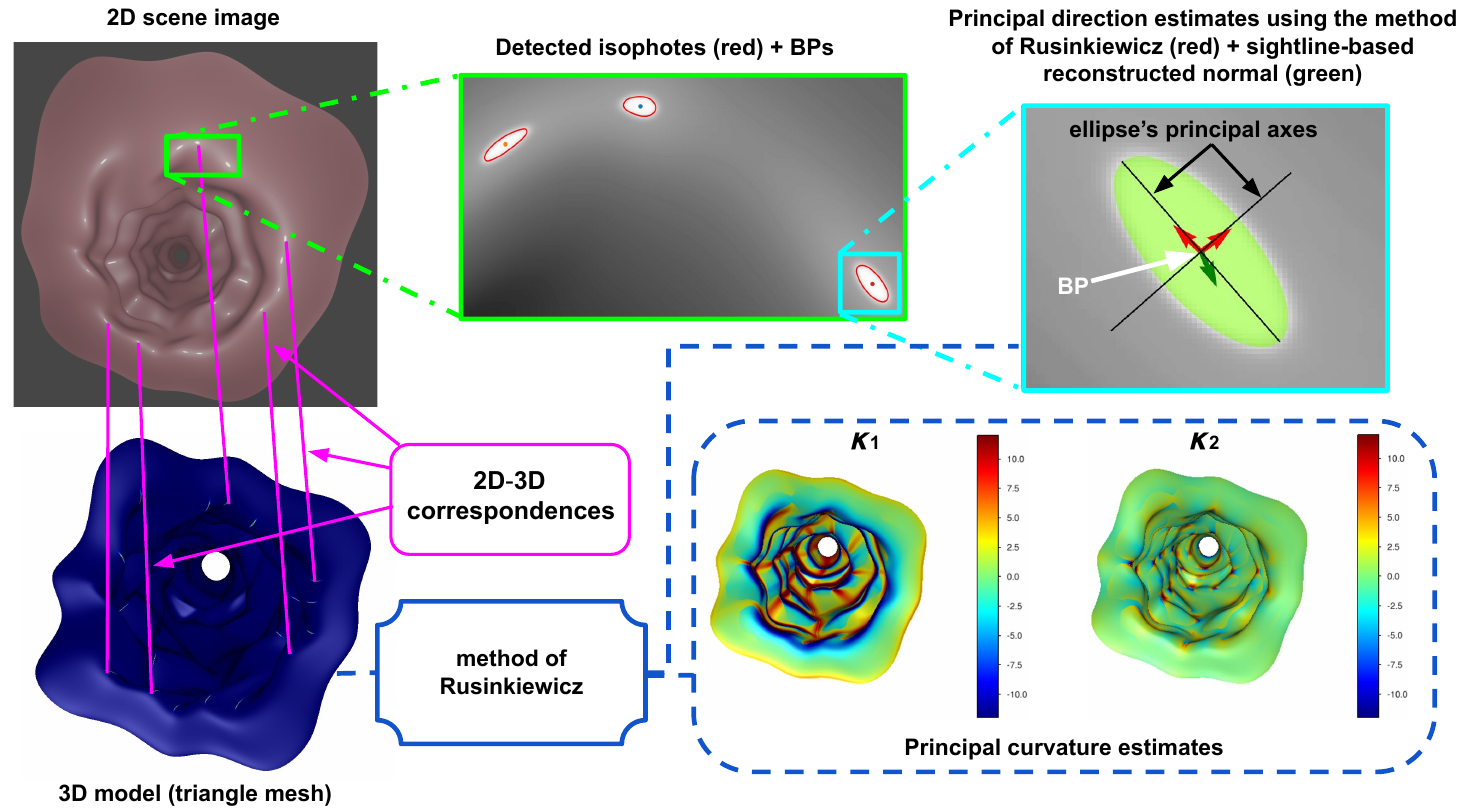}
\caption{\label{fig:fig2} Colon-like synthetic sequence. The available 2D-3D correspondences (in magenta) are used to compare between our reconstruction of the normal, the principal curvature ratio and the principal directions for the surface at specularity's brightest point and the same quantities estimated from the 3D model via the method of Rusinkiewicz~\cite{rusinkiewicz2004estimating}.}
\end{figure}

\begin{figure}[h!]
    \centering
    \begin{minipage}{0.235\textwidth}
        \centering
        \includegraphics[width=0.9\textwidth]{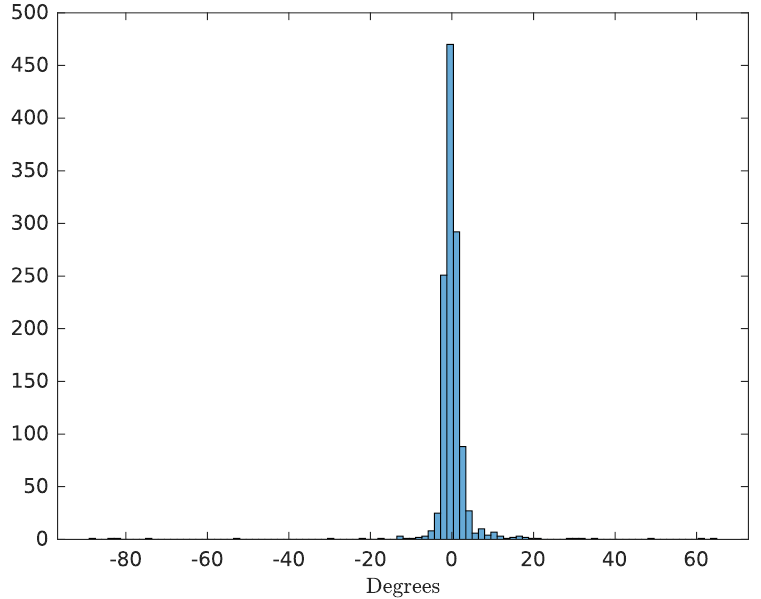} 
        \caption{Error histogram: angle between reconstructed and true normals (in degrees).} \label{fig:3}
    \end{minipage}\hfill
    \begin{minipage}{0.235\textwidth}
        \centering
        \includegraphics[width=0.9\textwidth]{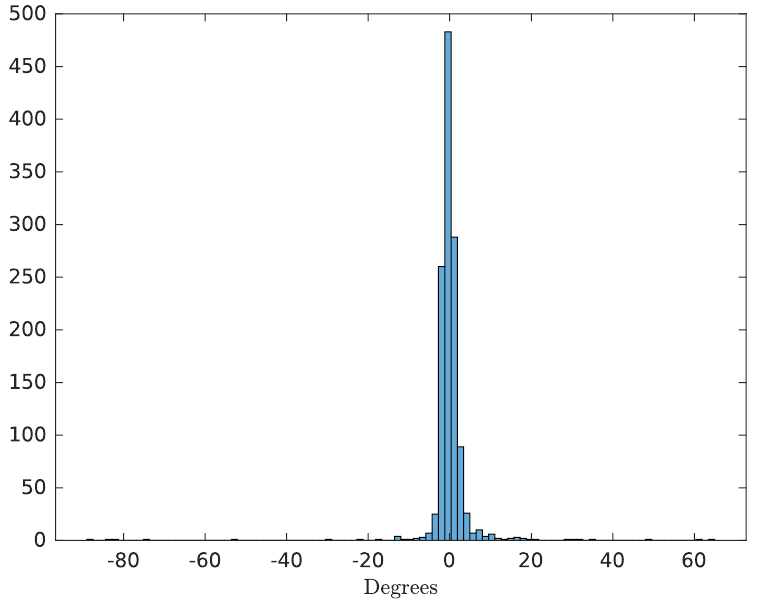} 
        \caption{angle between reconstructed normals and $ \vec{v_1} \times \vec{v_2}$ (in degrees).}
        \label{fig:4}
    \end{minipage}
\end{figure}

We estimate the true principal curvatures $\kappa_1$ and $\kappa_2$ as the eigenvalues of the second fundamental form that we compute numerically by the method~\cite{rusinkiewicz2004estimating}, from which we obtain the true curvature ratio.
We also compute the true principal directions, as the eigenvectors of the second fundamental form.
A representative example is shown in figure~\ref{fig:fig2} (top right), showing that the specular ellipse's axes match the principal directions.
We perform a sanity check by verifying that the reconstructed normal is orthogonal to the reconstructed principal directions on the tangent plane, confirming that they together form an orthonormal local coordinate system.
The histogram of errors in principal direction estimates is shown in figure~\ref{fig:4}.

\subsection{Real Data}
\label{subsec:real}
\begin{figure}[!h]
\centering
\includegraphics[width=0.5\textwidth]{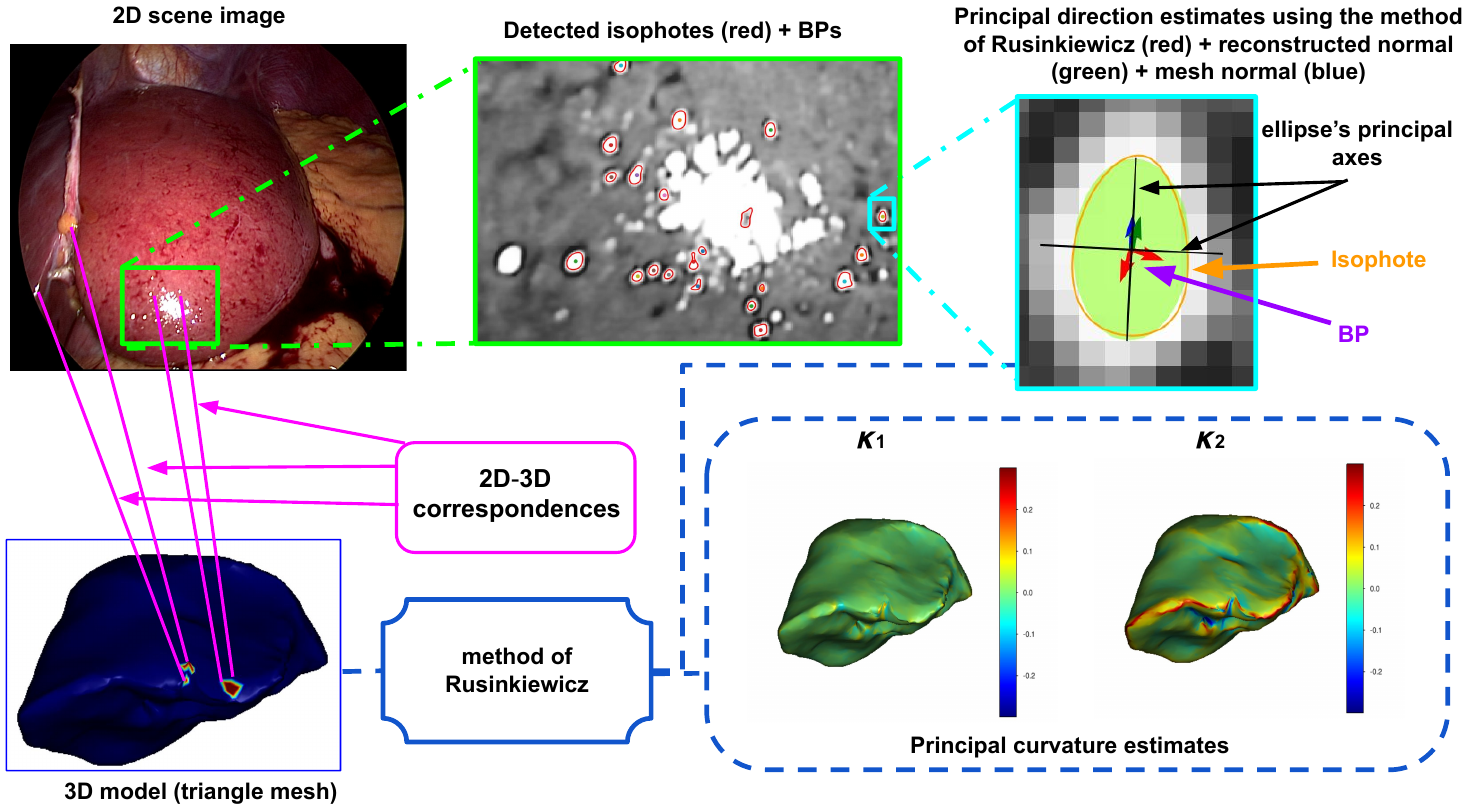}
\caption{\label{fig:liver} Qualitative results for the liver data.}
\end{figure}

The applicability of our reconstruction approach to real data is evaluated in 2D laparoscopic views of the liver, where a sparse set of specular highlights is detected and for which we have 3D \textit{reference models}.
These references models are obtained by 3D-to-2D registration of a virtual patient model segmented from a preoperative CT scan by experts.
The registration is solved by the method from~\cite{espinel2022using} to a satisfying accuracy.
We use three images extracted from three procedures collected in our hospital under ethical approval IRB00008526-2019-CE58 issued by CPP Sud-Est VI in Clermont-Ferrand, France. A representative example is shown in figure~\ref{fig:liver}.

\subsubsection{Normal Reconstruction}
\label{subsubsec:Normal reconstruction2}
\begin{figure}[!h]
\centering
\includegraphics[width=0.4\textwidth]{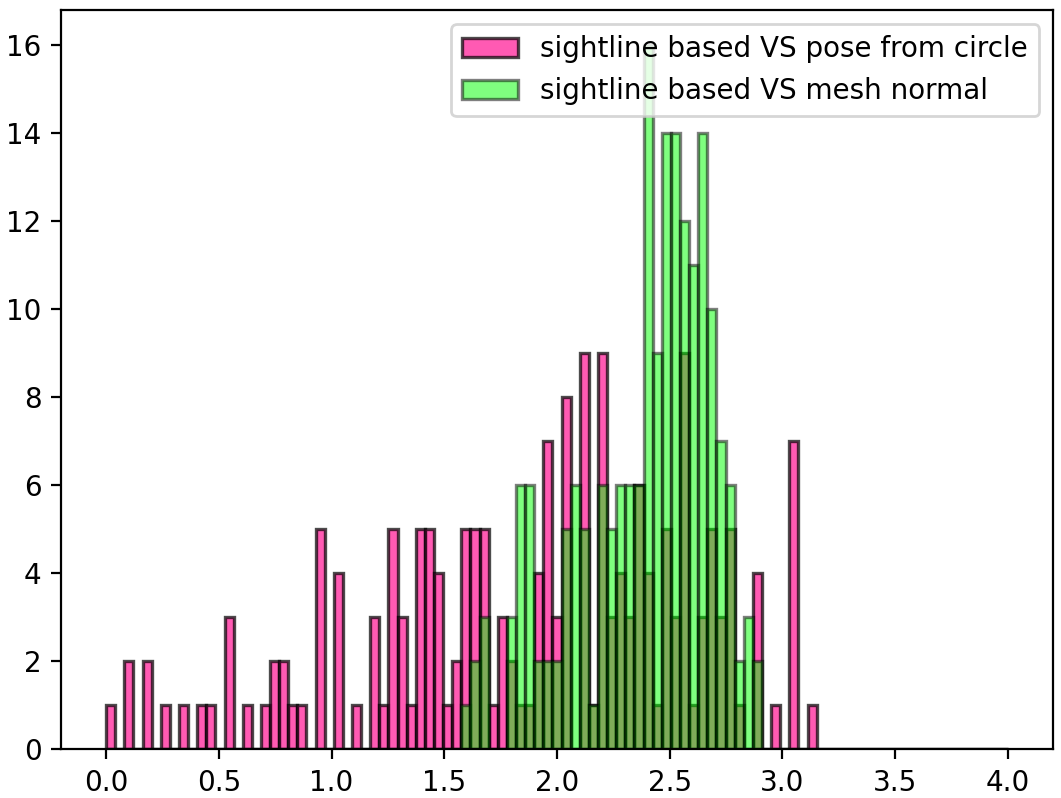}
\caption{\label{fig:histogram_liver} Evaluation of normal reconstruction for the laparoscopic liver images. The histograms show the absolute angular difference (in degrees).}
\end{figure}

Contrary to~\cite{makki2023normal} in which a small set of isolated specularities is manually selected to then detect isophotes using the marching squares, we automatically detect numerous specular isophotes from our neural network~\cite{makki2023elliptical}. An example showing elliptical isophotes is shown in figure~\ref{fig:liver}. 193 small elliptical specularities are kept from the three selected images according to the ellipticity  criterion and used to compare the reconstructed normals with two other estimates: 1) the normals produced by the pose from circle method~\cite{makki2023normal}, and 2) the normals directly determined from the registered virtual 3D model. The histograms of figure~\ref{fig:histogram_liver} show a good agreement between the three methods. The green histogram shows an angular error centred at 2.5$^{\circ}$, which confirms the efficiency of both the proposed and the registration methods~\cite{koo2017deformable}. The improvement is partly due to the ellipticity, which is not present in~\cite{makki2023elliptical}. Nevertheless, the magenta histogram shows an angular error centred at 3$^{\circ}$, which shows a fair agreement between the proposed and the previous methods.

\subsubsection{Principal directions principal curvature ratio }
\label{subsubsec:principal directions2}

\begin{figure}[!h]
    \centering
    \subfloat[\centering principal directions (differences $\left | \theta_e \right |$ in degrees)]{{\includegraphics[width=3.85cm]{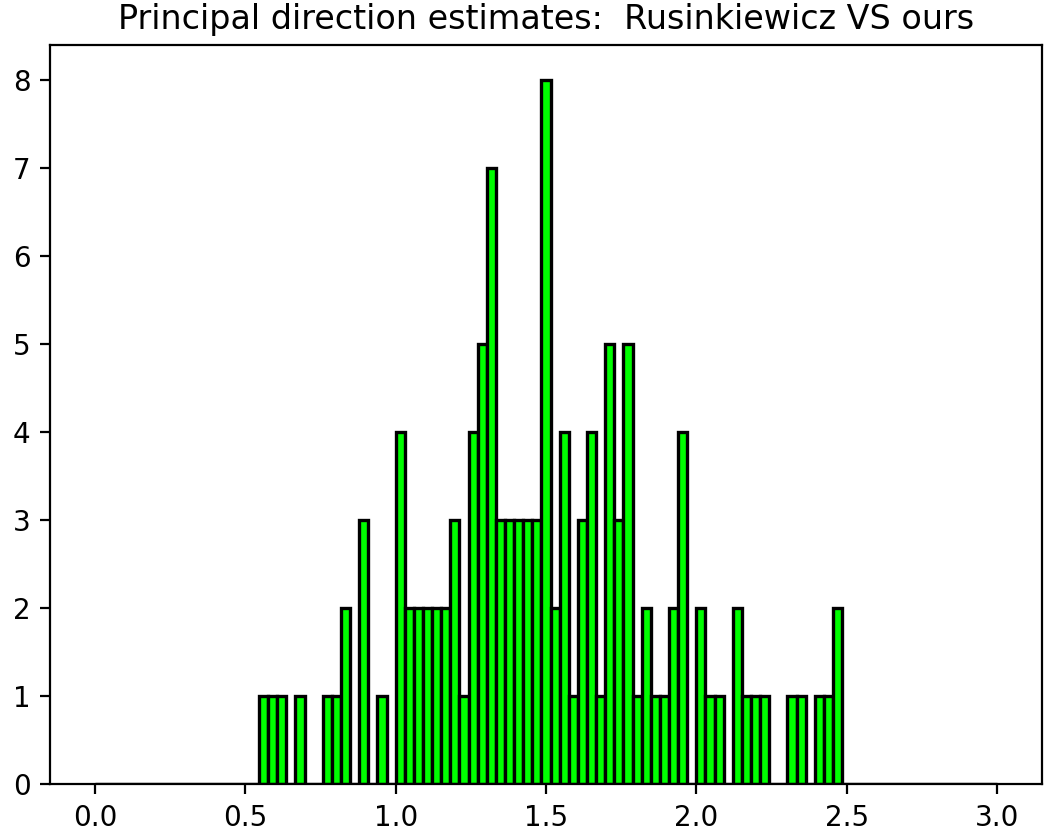} }}%
    \qquad
    \subfloat[\centering principal curvature ratio (differences $d$ in a.u.)]{{\includegraphics[width=3.85cm]{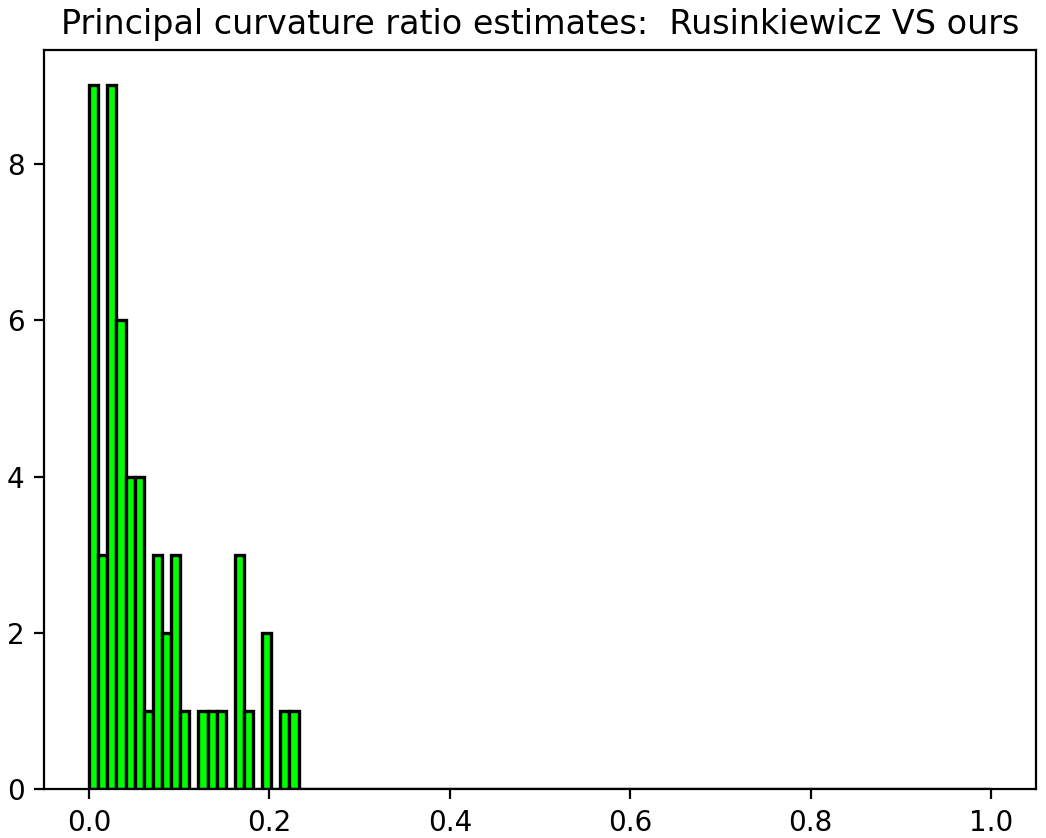} }}%
    \caption{Evaluation of shape reconstruction: direct comparison between our method and~\cite{rusinkiewicz2004estimating}.}%
    \label{fig:liver3.2.2}%
\end{figure}

First, in order to evaluate the quality of our reconstructed principal directions $\{ \vec{v_1} , \vec{v_2}\}$, we have compared them to the principal directions $\{ \mathbf{u}_1 , \mathbf{u}_2\}$ estimated using~\cite{rusinkiewicz2004estimating} from the triangle mesh.  An easy way to do that is to quantify the angle between their cross products since both of them are supposed to be aligned with the normal direction.
Let $\theta_e = \angle (\mathbf{u}_1 \times \mathbf{u}_2, \vec{v_1} \times \vec{v_2}) $ be this angle expressed in degrees. We then consider its absolute value $\left | \theta_e \right |$ to quantify the degree of agreement between the two methods, leading to the histogram error illustrated in figure~\ref{fig:liver3.2.2}(a). This histogram shows a small error which is always less than 3$^{\circ}$ for the same specularities used in Section~\ref{subsubsec:Normal reconstruction2} for normal reconstruction evaluation.

Second, we evaluated the differences between the two estimated curvature ratios (proportional to the eccentricity) as $d = \left | \left |\lambda_1 /\lambda_2 \right | - \left |\kappa_1 /\kappa_2 \right | \right |$ where $\kappa_1$ and $\kappa_2$ are the minimum and maximum curvatures estimated using~\cite{rusinkiewicz2004estimating}. The histogram of figure~\ref{fig:liver3.2.2}(b) shows that the error was almost comprised between 0 and 0.2 (and mostly close to zero) which is still very acceptable since the eccentricity of an ellipse always lies between 0 (a perfect circle) and 1 (a straight line).  The example presented in figure~\ref{fig:liver} (top right) show that the reconstructed normals and principal direction are of good quality.

We finally show qualitative results for a colonoscopic image extracted from the EndoMapper database~\cite{azagra2023endomapper} and for which high quality 3D models are not available. 
\begin{figure}[!h]
\centering
\includegraphics[width=0.5\textwidth]{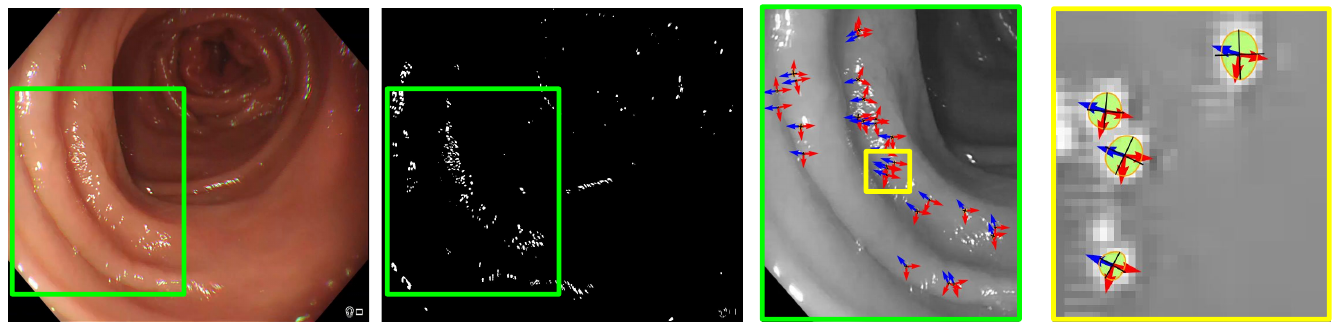}
\caption{\label{fig:colon} Colonoscopy. From left to right: image $I$, neural mask $M$, the third and fourth images show some reconstructed principal directions (red) and normals (blue).}
\end{figure}

\section{Conclusion}
\label{sec:conclusion}
We have presented a mathematical model based on a second-order model of the tissue surface at specularities. 
This model leads to a reconstruction method which can infer the local surface normal, principal directions and curvature ratio.
The method is simple and fast. It produces normal estimates with accuracy on par with previous methods, solves the local two-fold ambiguity and estimates an additional level of local shape information.
As it works from a single image, we plan to use it to boost the NRSfM method in the endoscopic case.

\bibliographystyle{IEEEbib}
\bibliography{strings,refs}

\end{document}